\pdfoutput=1
\documentclass[11pt]{article}

\usepackage{EMNLP2023}

\usepackage{times}
\usepackage{latexsym}
\usepackage[T1]{fontenc}
\usepackage[utf8]{inputenc}
\usepackage{microtype}
\usepackage{inconsolata}
\usepackage{xspace}
\usepackage{amsmath}
\usepackage{graphicx}
\usepackage{xurl}
\usepackage{array}

\newcommand{\warpv}{WARP\textsubscript{V}\xspace}
\newcommand{\amulap}{AMuLaP\xspace}
\newcommand{\laav}{LAAV\xspace}
\newcommand{\mask}{\texttt{[MASK]}\xspace}
\newcommand{\V}{\ensuremath{\mathcal{V}_M}\xspace}
\newcommand{\VS}[1]{\ensuremath{\mathcal{S}(#1)}\xspace}

\title{Label-Aware Automatic Verbalizer for Few-Shot Text Classification}

\author{Thanakorn Thaminkaew$^1$, Piyawat Lertvittayakumjorn$^2$, Peerapon Vateekul$^1$\thanks{\quad Corresponding author}\\
  $^1$ Department of Computer Engineering, Faculty of Engineering, \\Chulalongkorn University, Thailand \\
  $^2$ Google, United States\\
  \texttt{6472031921@student.chula.ac.th, piyawat@google.com, peerapon.v@chula.ac.th}}

\begin{document}
\maketitle

\begin{abstract}
Prompt-based learning has shown its effectiveness in few-shot text classification. One important factor in its success is a verbalizer, which translates output from a language model into a predicted class. Notably, the simplest and widely acknowledged verbalizer employs manual labels to represent the classes. However, manual selection does not guarantee the optimality of the selected words when conditioned on the chosen language model. Therefore, we propose Label-Aware Automatic Verbalizer (\laav), effectively augmenting the manual labels to achieve better few-shot classification results. Specifically, we use the manual labels along with the conjunction "and" to induce the model to generate more effective words for the verbalizer. The experimental results on five datasets across five languages demonstrate that \laav significantly outperforms existing verbalizers. Furthermore, our analysis reveals that
\laav suggests more relevant words compared to similar approaches, especially in mid-to-low resource languages.
\end{abstract}

\section{Introduction} \label{introduction}
In recent years, we have seen many promising applications of \emph{prompt-based learning} for text classification \cite{schick-schutze-2021-just,wang-etal-2022-towards-unified,zhang-etal-2022-prompt-based,hu-etal-2022-knowledgeable}.
While the traditional approach trains or fine-tunes a model to directly predict a class for an input text, the prompt-based approach fits the input text into a \emph{template} that has some slots to be filled. Next, it asks a language model (LM)\footnote{Generally, masked LMs are preferred for classification tasks because they require fewer resources to fine-tune \cite{liu2023pre}.} to fill in the slots and then translate what the model fills to be a predicted class \cite{liu2023pre}.
To predict whether a movie review "Great movie!" has a positive or negative sentiment, we may prompt a masked LM with "Great movie! It was \mask." The model may predict the word "fun" for the \mask token, and we can apply a function, so-called a \emph{verbalizer}, to map "fun" to the positive class.

\begin{figure}[t!]
    \centering
    \includegraphics[width=7.7cm]{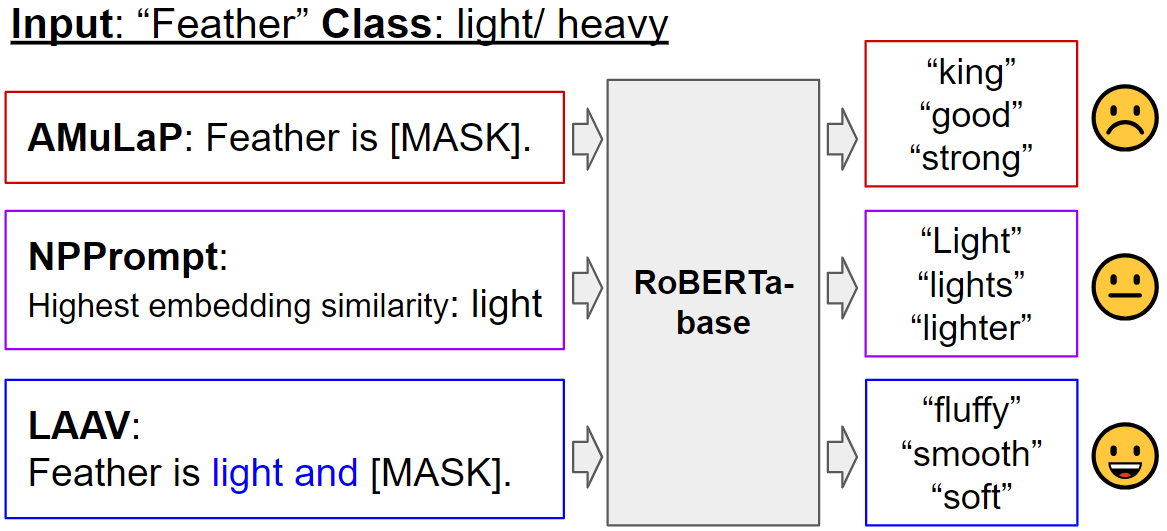}
    \caption{Illustration of \laav compared to \amulap and NPPrompt when searching for class representative tokens.}
    \label{fig:overview}
\end{figure}

Certainly, one important factor that defines the success of a prompt-based text classifier is its verbalizer. \citet{schick-schutze-2021-exploiting} proposed PET, which manually chooses a word to represent each class. During inference, it compares the likelihood of those words at the \mask token (as predicted by the LM) to find the most probable class. In contrast, \citet{wang-etal-2022-automatic} proposed \amulap, which represents each class with a set of words, automatically derived from those predicted by the LM for training examples.
However, there is no guarantee that the words chosen by the LM will be relevant to the classes of interest. \citet{zhao-etal-2023-pre} proposed NPPrompt, which represents each class using a set of tokens with the highest embedding similarity to the manual class label. Its performance, therefore, relies solely on the LM’s embedding space.

As illustrated in Figure~\ref{fig:overview} (top), to predict whether an object "Feather" is light with a prompt "Feather is \mask.", the LM suggests "king", "good", and "strong", which are irrelevant to the task
but used by \amulap to construct the verbalizer.
Meanwhile, as shown in Figure~\ref{fig:overview} (middle),
NPPrompt suggests "Light", "lights", and "lighter", which are variations related to the class "light" but hardly provide additional information about the class.

In this paper, we propose \textbf{\laav} (\textbf{L}abel-\textbf{A}ware \textbf{A}utomatic \textbf{V}erbalizer), which improves \amulap by exploiting the class labels to induce the model to generate more relevant words for the verbalizer. As shown in Figure \ref{fig:overview} (bottom), we could construct a better verbalizer by asking "Feather is \textbf{light and} \mask." Now, the LM suggests "fluffy", "smooth", and "soft", which are closely connected to the light class. Overall, the contributions of this paper are as follows.

\begin{itemize}
\setlength\itemsep{-0.1em}
    \item We propose \laav -- a simple yet effective technique to create a reliable verbalizer for prompt-based text classification (Section~\ref{sec:laav}).
    
    \item We conduct few-shot classification experiments on five datasets from five languages (Section~\ref{sec:experiments}). The results show that \laav outperforms the baselines in most settings, especially when we have 16 training examples per class or fewer (Section~\ref{subsec:mainresult}). 
    
    \item We carry out additional analyses to determine the best choice of conjunction for retrieving more related words (Section~\ref{subsec:conjunction}) and to investigate the quality of the words selected by \laav compared to \amulap (Section~\ref{subsec:verbalizerinterpret}).
\end{itemize}

\section{Background \& Related Work} \label{sec:relatedwork}
\subsection{Few-shot Text Classification}
In text classification, various strategies have been employed for addressing few-shot scenarios. 
For example, meta-learning utilizes labeled examples from auxiliary tasks to train a model that can quickly adapt to new tasks with only a few examples \cite{li2020few, yin2020meta}.
The semi-supervised or weakly-supervised approach leverages extensive unlabeled data in conjunction with the limited labeled data to improve the model's performance \cite{li2018law, duarte2023review}.
In-context learning includes a few labeled examples as demonstrations in a prompt for querying large PLMs to get the classification \cite{brown2020language, lin2021few}.
Our paper, in contrast, follows the prompt-based learning approach involving template design, verbalizer, and model fine-tuning.
This approach has shown promising results in enhancing model training efficiency in several previous works \cite{zhao-etal-2023-pre,schick-schutze-2021-exploiting,wang-etal-2022-automatic}.
Also, it has benefits over other approaches in situations where the auxiliary tasks, the unlabeled data, and the large PLMs are hardly available (e.g., few-shot classification in mid-to-low resource languages explored in this paper).

\subsection{Verbalizers for Prompt-Based Learning}\label{verbalizerconstruction}
The easiest way to construct a verbalizer is to manually select a representative word for each class, as in PET \cite{schick-schutze-2021-exploiting}. However, manual selection could be laborious (especially for multi-class classification) and does not guarantee the optimality of the selected words when conditioned on the chosen LM.
To automate verbalizer construction, \citet{hambardzumyan-etal-2021-warp} introduced trainable continuous tokens to serve as class representations, known as a soft verbalizer. Nonetheless, the obtained tokens may not correspond to actual words, hindering model debugging and improvement.
Meanwhile, some other works, including ours, still opt for discrete verbalizers, which provide more interpretability.
\citet{schick-etal-2020-automatically} searched for the best word to represent each class by maximizing the likelihood of the training data. \amulap \cite{wang-etal-2022-automatic} does the same but represents each class by multiple words instead to reduce the effects of noise in the data. 
NPPrompt \cite{zhao-etal-2023-pre} utilizes a set of tokens that have the closest embedding similarity to the manual label to represent each class. However, its effectiveness is strongly dependent on the quality of the LM's embedding space, which may not be effective for mid-to-low resource languages. Additionally, this approach neglects the input text, potentially causing issues with polysemous words that have multiple meanings.
Since our work is based on \amulap, the next section explains \amulap in more detail.  

\subsection{\amulap}
For a text classification task aiming to classify an input text $x$ to a class $y \in Y$, \amulap represents each class $y_i$ with a set of $k$ tokens, denoted as \VS{y_i}. These tokens are selected from the sub-word vocabulary \V of the language model $M$ it prompts. To construct \VS{y_i}, it applies a template $T$ to all training examples $x$ of which the ground truth label is $y_i$. One example is $T(x) = \: [x] \: It \: was \: \mask$ for the sentiment analysis task in the Introduction. Then it lets $M$ predict the probability of each $v \in \V$ for the \mask of these $T(x)$s. The score of token $v$ for class $y_i$ is

\begin{equation}
s(v, y_i) = \sum_{(x, y_i) \in D} p_M( \mask = v|T(x))
\end{equation}

where $D$ is the training set and $p_M$ is the probability predicted by $M$. \VS{y_i} is then defined as a set of $k$ tokens with the highest $s(v, y_i)$. 
To ensure that each token $v$ is assigned to only one class, \amulap calculates its score for every $y \in Y$ and assigns it to the class $y_i$ where $y_i = \arg \max_{y \in Y} s(v, y)$.

After that, the LM is fine-tuned on $D$ using the cross-entropy loss. Specifically, the log-probability of class $y_i$ for an input $x$ is 

\begin{equation}
\label{eqn:logprobabilityscore}
L(y_i|x) = \frac{1}{k} \sum_{v \in \VS{y_i}} \log p_M(\mask = v |T(x))
\end{equation}

The cross-entropy loss will be calculated from $L(y_i|x)$ for all $y_i \in Y$ and all $x \in D$ as

\begin{equation}
loss = -\sum_{(x,y) \in D} \sum_{y_i \in Y} I(y,y_i) \cdot L(y|x)
\end{equation}
where $I(y,y_i) = 1$ if $y=y_i$; otherwise, 0.

Finally, during validation and testing, the predicted label $\hat{y}$ for an input $x$ is simply $\arg \max_{y_i \in Y} L(y_i|x)$.

\section{Label-Aware Automatic Verbalizer} \label{sec:laav}

As explained in Section \ref{introduction}, the words in \VS{y_i}, selected by \amulap, could be unrelated to their corresponding class. So, when constructing \VS{y_i}, our method \laav integrates the label name of $y_i$ into the template $T$,  using a conjunction. This helps induce $M$ to predict words that are related to $y_i$. Our choice for the conjunction is "and" because it serves to connect words or phrases with the same grammatical category and similar meaning. Also, "and" is one of the most widely used conjunctions in many languages \cite{davies2011word}. As a result, our \laav template for creating \VS{y_i} is

\begin{equation*}
    T_{y_i}(x) = \: [x] \: It \: was \: [y_i] \: \textbf{and}  \: \mask
\end{equation*}

Note that we will explore other conjunction options in Section~\ref{subsec:conjunction}.
Now, the score of token $v$ for class $y_i$ for \laav will be

\begin{equation}
s(v, y_i) = \sum_{(x, y_i) \in D} p_M( \mask = v|T_{y_i}(x))
\end{equation}

Since the objective of the \laav template $T_{y_i}$ is solely for seeking better representative words for each class, we use the original template $T$ without the conjunction during training and inference. 

\section{Experiments} \label{sec:experiments}
\subsection{Datasets and Pre-trained Models}

We conducted experiments on five datasets from five languages. 
These include AG’s News (English) \cite{zhang2015character}, which is a news classification dataset, and the other four sentiment analysis datasets, i.e., SmSA (Indonesian) \cite{wilie-etal-2020-indonlu}, Students' Feedback (Vietnamese) \cite{van2018uit}, Wisesight sentiment (Thai) \cite{https://doi.org/10.5281/zenodo.3457447}, and Shopee Reviews (Tagalog) \cite{neil_riego_2023}. The \laav templates, the class labels, and other details of each dataset are reported in Appendix \ref{sec:datasetdetails}.

The pre-trained LMs used in this paper are the -base versions of RoBERTa \cite{liu2019roberta}, IndoBERT \cite{wilie2020indonlu}, Tagalog RoBERTa \cite{cruz2021improving}, WangchanBERTa \cite{lowphansirikul2021wangchanberta}, and PhoBERT \cite{phobert} for English, Indonesian, Tagalog, Thai, and Vietnamese, respectively.

\subsection{Implementation Details}
In a few-shot scenario, we randomly selected 4, 8, 16, or 32 samples per class for both the training and validation splits. 
Since we do not have a sizable development set for optimizing hyperparameters, we depend on related work to guide us in selecting the appropriate hyperparameters. 
All text inputs were limited to 500 characters. During training, we used Adam optimizer \cite{kingma2014adam} with a learning rate of 1e-5 to optimize the loss function. 
To prevent overfitting, we employed an early stopping method with a maximum of 100 epochs. We repeated the training process five times using different seeds to ensure robustness. We set $k$ = 32 for all experiments. Our models were implemented using PyTorch \cite{paszke2019pytorch}, Huggingface transformers \cite{wolf2019huggingface}, and the OpenPrompt \cite{ding2021openprompt} libraries and trained on a Tesla P100 PCIe 16 GB. 

\subsection{Baselines}
We evaluated our method by comparing it to 
\textbf{Traditional Fine-tuning} (i.e., plugging a linear classification layer of top of the [CLS] embedding of the LM and fine-tuning the whole model) 
and four recent verbalizer methods including 
(1) \textbf{PET} manually selecting a token to represent each class \cite{schick-schutze-2021-exploiting}, 
(2) the verbalizer of WARP, denoted as \textbf{\warpv}, representing each class with a trained continuous vector \cite{hambardzumyan-etal-2021-warp}, 
(3) \textbf{PETAL} searching for the most suitable representative token \cite{schick-etal-2020-automatically}, and 
(4) \textbf{\amulap} searching for multiple suitable representative tokens using an unmodified template \cite{wang-etal-2022-automatic}.
(5) \textbf{NPPrompt} searching for multiple suitable representative tokens using a set of tokens with the highest embedding similarity to the manual label \cite{zhao-etal-2023-pre}.
We utilized the OpenPrompt \cite{ding2021openprompt} library to implement \warpv (using SoftVerbalizer) and PETAL (using AutomaticVerbalizer). For the other baselines, we implemented them manually in Pytorch \cite{paszke2019pytorch}.

\section{Results and Additional Analyses}
\subsection{Comparison to the Baselines} \label{subsec:mainresult}
Table~\ref{tab:mainresult} shows the results of our method compared to the baselines. 
Note that we cannot apply PET to the Shopee Reviews dataset (Tagalog) because the label "napakasama" (very bad) cannot be presented using a single token in Tagalog RoBERTa. 

Overall, our method, \laav, works well in few-shot learning, especially with 16 training examples per class or fewer. In an extreme few-shot setting with only 4 training examples per class, our model improves the Macro F1 scores by at least 1.2\% absolute from other baselines across five datasets. This shows that the representative tokens selected by \laav result in greater performance. 
When the number of training examples increases to 32, however, Traditional Fine-tuning performs on par or better compared to the prompt-based methods including \laav in several datasets due to the sufficient number of training examples the LMs can effectively learn from. 

\begin{table}[t]
\centering
\setlength\extrarowheight{1pt}
\small
\resizebox{7.7cm}{!}{
\begin{tabular}{lcccc}
\hline
Sample Size & 4           & 8           & 16          & 32          \\ \hline
\multicolumn{5}{l}{AG’s News (English)}                             \\ \hline
Traditional FT        & 75.6 (4.9)  & 81.7 (4.9)  & 84.8 (5.8)  & 87.0 (5.8)  \\
PET         & 79.1 (5.1)  & 83.8 (5.9)  & 85.9 (6.1)  & 86.7 (4.7)  \\
WARP\textsubscript{V}        & 70.4 (5.6)  & 75.4 (5.3)  & 81.9 (6.7)  & 84.7 (6.4)  \\
PETAL       & 68.1 (7.2)  & 79.0 (8.6)  & 82.2 (9.9)  & 85.4 (10.1) \\
AMuLaP      & 71.6 (5.9)  & 78.3 (6.6)  & 83.3 (6.9)  & 86.2 (7.3)  \\
NPPrompt    & 79.9 (2.1)  & 82.7 (3.0)  & 84.2 (2.6)  & 86.0 (2.6)  \\
Ours: LAAV        & \textbf{81.1 (1.2)}  & \textbf{84.2 (2.0)}  & \textbf{86.6 (2.3)}  & \textbf{87.7 (2.2)}  \\ \hline
\multicolumn{5}{l}{SmSA (Indonesian)}                               \\ \hline
Traditional FT        & 48.1 (7.4)  & 52.2 (9.6)  & 62.3 (8.9)  & 67.4 (8.0)  \\
PET         & 49.1 (8.4)  & 53.0 (9.2)  & 63.9 (8.9)  & 70.2 (7.5)  \\
WARP\textsubscript{V}        & 50.9 (7.2)  & 52.2 (7.6)  & 59.7 (6.2)  & 67.7 (4.4)  \\
PETAL       & 53.8 (6.2)  & 52.1 (6.6)  & 65.1 (6.5)  & 69.5 (6.9)  \\
AMuLaP      & 58.9 (4.6)  & 58.3 (5.2)  & 64.7 (4.9)  & 69.5 (4.6)  \\
NPPrompt    & 50.7 (6.4)  & 51.6 (6.2)  & 62.0 (7.0)  & 69.8 (7.0)  \\
Ours: LAAV        & \textbf{61.1 (7.6)}  & \textbf{58.5 (7.8)}  & \textbf{70.1 (7.7)}  & \textbf{73.2 (6.1)}  \\ \hline
\multicolumn{5}{l}{Shopee Reviews (Tagalog)}                        \\ \hline
Traditional FT        & 27.4 (3.8)  & 31.1 (6.6)  & 34.9 (6.5)  & \textbf{38.8 (6.5)}  \\
PET         & -           & -           & -           & -           \\
WARP\textsubscript{V}        & 25.1 (2.1)  & 28.1 (4.0)  & 31.1 (3.9)  & 34.4 (3.9)  \\
PETAL       & 26.8 (3.9)  & 30.2 (4.6)  & 33.2 (4.2)  & 35.0 (4.1)  \\
AMuLaP      & 28.9 (5.8)  & 32.4 (6.1)  & 35.4 (6.2)  & 37.2 (6.2)  \\
NPPrompt    & 17.9 (7.4)  & 26.9 (10.4) & 27.6 (10.4) & 31.0 (10.9) \\
Ours: LAAV        & \textbf{31.6 (3.7)}  & \textbf{32.6 (3.7)}  & \textbf{36.1 (3.6)}  & 36.1 (3.7)  \\ \hline
\multicolumn{5}{l}{Wisesight sentiment (Thai)}                      \\ \hline
Traditional FT        & 28.2 (4.2)  & 29.6 (3.2)  & 38 (3.4)    & 42.1 (4.7)  \\
PET         & 34.5 (6.5)  & 41.0 (6.4)  & 47.2 (6.4)  & 47.9 (7.1)  \\
WARP\textsubscript{V}        & 30.8 (4.3)  & 37.7 (4.3)  & 39.4 (5.2)  & 42.9 (5.7)  \\
PETAL       & 30.8 (4.4)  & 37.1 (4.7)  & 44.2 (6.4)  & 44.8 (3.5)  \\
AMuLaP      & 32.3 (5.6)  & 37.4 (4.6)  & 45.5 (5.7)  & 48.7 (4.7)  \\
NPPrompt    & 31.0 (7.8)  & 37.0 (9.4)  & 43.6 (5.0)  & 45.3 (3.9)  \\
Ours: LAAV        & \textbf{38.1 (4.5)}  & \textbf{42.1 (4.4)}  & \textbf{48.5 (2.8)}  & \textbf{51.2 (2.9)}  \\ \hline
\multicolumn{5}{l}{Students' Feedback (Vietnamese)}                 \\ \hline
Traditional FT        & 52.0 (9.9)  & 62.6 (11.4) & 71.6 (11.9) & 73.4 (12.2) \\
PET         & 65.5 (3.0)  & 68.7 (2.9)  & 70.3 (3.9)  & \textbf{74.1 (4.7)}  \\
WARP\textsubscript{V}        & 51.3 (8.3)  & 57.3 (9.0)  & 63.1 (9.4)  & 65.1 (9.4)  \\
PETAL       & 49.1 (8.9)  & 57.7 (11.3) & 66.3 (11.2) & 66.6 (11.2) \\
AMuLaP      & 55.6 (11.2) & 64.6 (11.3) & 71.3 (11.7) & 72.6 (12.0) \\
NPPrompt    & 38.5 (15.8) & 40.0 (15.8) & 58.7 (12.2) & 49.1 (12.2) \\
Ours: LAAV        & \textbf{67.9 (2.8)}  & \textbf{69.5 (2.9)}  & \textbf{71.7 (3.5)}  & 72.7 (3.1)  \\ \hline
\end{tabular}
}
\caption{Macro F1 results along with their standard
deviation in the parentheses tested on five datasets. The best results are marked in \textbf{bold}.} \label{tab:mainresult}
\end{table}

\begin{table}[t]
\centering
\resizebox{7.7cm}{!}{
\begin{tabular}{l|lc|cc}
\hline
Dataset & Translated Result & Search & "of" & "and" \\ \hline
AG's News & \underline{and}, to, for & \textbf{84.88} & 84.19 & \textbf{84.88} \\
SmSA & \underline{exchange}, dough, mop & 63.58 & 64.33 & \textbf{65.71} \\
Shopee Reviews & \underline{in}, that, of & 30.55 & 32.44 & \textbf{34.11} \\
Wisesight sentiment & \underline{more}, of, that & 38.57 & 43.29 & \textbf{44.97} \\
Students' Feedback & \underline{of}, give, and & 61.84 & 62.14 & \textbf{70.45} \\ \hline
\end{tabular}
}
\caption{Macro F1 results of candidate conjunctions, averaged from the 4, 8, 16, 32-shot settings. The best English-translated tokens from the search process are \underline{underlined}, with their results reported in the "Search" column.} 
\label{tab:conjsearch}
\end{table}

\subsection{Choices of conjunction} \label{subsec:conjunction}
While we used "and" as the conjunction of \laav templates so far, this section aims to explore whether there are other promising conjunction choices we missed. 
Hence, we designed the following conjunction search process. 
First, we used \amulap to find the initial \VS{y_i} of each class. Then, we applied the template

\begin{equation*}
    T^S_{y_i}(x)= \: [x] \: It \: was \: [y_i] \: \mask \: [v]
\end{equation*}

for all $v \in \VS{y_i}$, to every training examples $x$ that has the label $y_i$. Basically, $T^S_{y_i}$ asks the LM to predict a token that can well connect $y_i$ to $v$, having the potential to be the conjunction in \laav template.

Table \ref{tab:conjsearch} shows the top three English-translated tokens (as averaged across the training sets) according to their language-specific LMs. 
"Of" and "and" appear most often across the languages, so we use them in \laav templates for every dataset and report their performance on the test set.
Furthermore, we try the best token (\underline{underlined}) of each dataset and report their performance under the "Search" column of Table \ref{tab:conjsearch}.
Ultimately, \textbf{"and"}  achieves consistently best results across datasets, supporting our initial \laav template design.

\subsection{Verbalizer Interpretation} \label{subsec:verbalizerinterpret}

To analyze the discriminative power of each token $v \in \VS{y_i}$,
we first split test examples into two groups: ones that belong to class $y_i$
and the others. For each group, we calculated the average logits (i.e., raw output from the LM) of the token $v$ using its training template $T(x)$ and the fine-tuned LM. Finally, we calculated the difference of average logits between the two groups. A token with high logits difference must receive significantly higher logits from examples of class $y_i$ than from those of the other classes, on average, showing its good quality to be used in the verbalizer.

\begin{figure}[t]
    \centering
    \includegraphics[width=7.7cm]{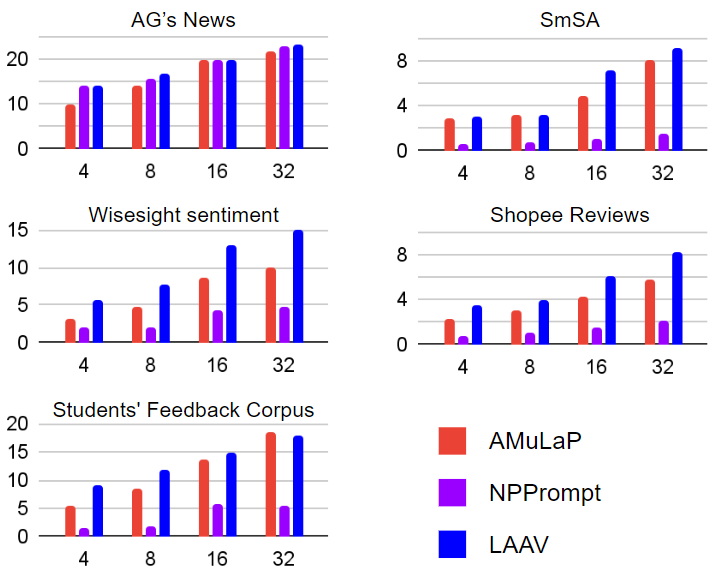}
    \caption{Comparison of the average logits difference across all words for each fine-tuned LM between \amulap, NPPrompt, and \laav.}
    \label{fig:logits}
\end{figure}

Based on Figure \ref{fig:logits}, the fine-tuned LMs using words from \laav consistently give higher logits difference compared to words from \amulap and NPPrompt, implying that words from \laav better represent the classes. 
This finding aligns with the accuracy data in Table \ref{tab:mainresult}, showing that a higher logits difference indicates increased discriminative power after fine-tuning the LM. For NPPrompt, the logits difference in the mid-to-low resource languages tends to be extremely low due to the limited vocabulary of the LM. This limitation results in the LM suggesting words related to the class name but associated with different parts of speech, ultimately hindering the fine-tuning process. For instance, in this experiment using the Thai LM (WangChanBERTa), the highest embedding similarity words to the class "question" are "doubt," "ask," and "answer," all of which are verbs whereas the \mask token in the template should be actually filled by a noun or an adjective.

We further examine the logits difference of each token. Table~\ref{tab:logitswords} compares words with the highest logits difference in AG's News from  \amulap and \laav. All top words from \laav are related to their respective class, while several top words from \amulap do not seem to relate to their class (e.g., "to" and "50"). On the other hand, NPPrompt suggests words that closely align with the class name itself, such as "Business" and "businesses" for the "business" class. These word suggestions restrict the diversity and depth of information, as they primarily rephrase the class name without delving into more contextually relevant aspects. This provides a reason for the higher performance of \laav in Section~\ref{subsec:mainresult}. 

\begin{table*}[t]
\centering
\resizebox{16cm}{!}{
\begin{tabular}{llll}
\hline
Class      & AMuLaP                                    & NPPrompt                                               & LAAV                                                 \\ \hline
world      & that, money, now, midnight, right         & world, World, WORLD, worlds, universe                      & history, politics, justice, time, home               \\
sports     & time, revenge, history, redemption, times & sports, Sports, sport, sporting, athletics                  & life, pride, culture, teamwork, winning              \\
business   & to, 50, 2020, 10, 40                      & business, Business, businesses, enterprise, corporate     & money, profit, people, family, entertainment         \\
technology & timing, speed, security, you, privacy     & technology, technologies, Technology, tech, technological & innovation, security, safety, reliability, community \\ \hline
\end{tabular}
}
\caption{Top-5 words with the highest logits difference on 4-shot settings to represent each class in AG's News.} \label{tab:logitswords}
\end{table*}

\section{Conclusion}
Our method, \laav, constructs a better verbalizer by exploiting class labels to collect more relevant words. As shown in the experiments, \laav outperforms other existing verbalizers in few-shot text classification across five languages. Our analyses show that "and" is a good conjunction to retrieve words that have high discriminative power for the classification task. In the future, we plan to explore the application of \laav in other scenarios such as multilingual LMs and multilabel classification.  

\section*{Limitations}

We only focused on improving the selection of words to represent each label with a fixed prompt template. Applying a tunable continuous template or a more specific discrete template may also reduce the ambiguity of the input and further improve the prompt-based learning results. 
In addition, with limited resources, we decided to explore experiments using the base version of the LMs. Fine-tuning larger LMs using parameter-efficient techniques may lead to different results. Nevertheless, the parameter-efficient techniques can be implemented on top of the prompt-based learning approach presented in this paper.

\appendix
\section{Dataset Details}
\label{sec:datasetdetails}

The dataset statistics, along with their respective \laav templates, \amulap templates, labels, and translated label names, are provided in the Table~\ref{tab:datasets}. Note that Shopee Reviews originally has five classes [1,..,5] which were manually mapped to textual labels ["very bad", ..., "excellent"]. All datasets are publicly available.
For languages other than English, we use Google Translate to construct their templates.

\begin{table}[h!]
    \centering
    \includegraphics[width=7.7cm]{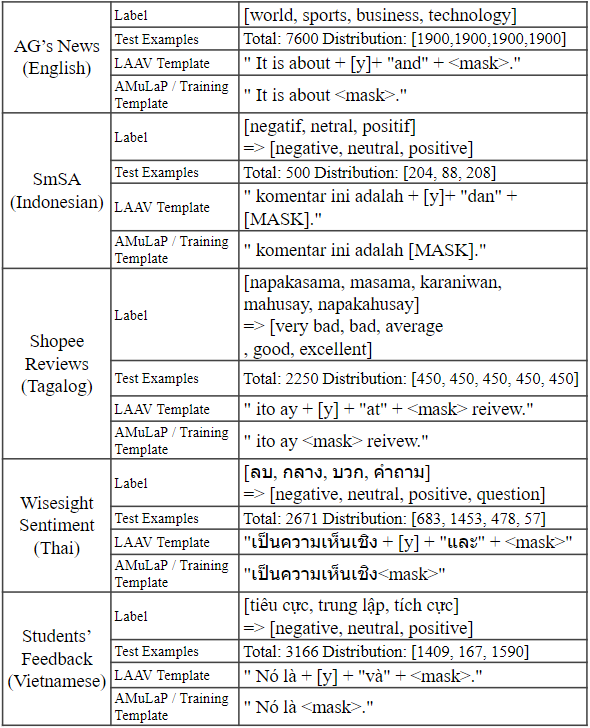}
    \caption{Details of the datasets along with their templates and labels.}
    \label{tab:datasets}
\end{table}

\end{document}